\documentclass{article}

\usepackage[final,main]{neurips_2026}

\usepackage[utf8]{inputenc}
\usepackage[T1]{fontenc}
\usepackage{hyperref}
\usepackage{url}
\usepackage{booktabs}
\usepackage{amsfonts}
\usepackage{amsmath}
\usepackage{amsthm}
\usepackage{nicefrac}
\usepackage{microtype}
\usepackage{xcolor}
\usepackage{graphicx}
\usepackage{wrapfig}
\usepackage[small]{caption}
\usepackage{algorithm}
\usepackage{algorithmic}
\usepackage{subcaption}
\usepackage{multirow}
\usepackage{colortbl}
\usepackage{xspace}
\usepackage[most]{tcolorbox}
\usepackage{newtxtt}

\graphicspath{{figures/}}

\newcommand{\model}{StaRPO\xspace}

\definecolor{best}{HTML}{B8D1E4}
\definecolor{second}{HTML}{E1F0F9}
\definecolor{bgcolor1}{HTML}{DAE8FC}
\definecolor{bgcolor2}{HTML}{D5E8D4}
\definecolor{bgcolor3}{HTML}{E1D5E7}
\definecolor{textred}{HTML}{B73F42}
\definecolor{textorange}{HTML}{DE9960}
\definecolor{starpoBoxBg}{HTML}{E1F0F9}
\definecolor{starpoBoxFrame}{HTML}{8EC3DC}

\tcbset{
  starpoOneBox/.style={
    enhanced,
    colback=starpoBoxBg,
    colframe=starpoBoxFrame,
    boxrule=0.6pt,
    arc=6pt,
    left=12pt,
    right=12pt,
    top=10pt,
    bottom=10pt,
    fontupper=\ttfamily\small,
  }
}

\title{StaRPO: Stability-Augmented Reinforcement Policy Optimization}

\author{%
  Jinghan Zhang$^{1}$,
  Fengran Mo$^{2}$,
  Tharindu Cyril Weerasooriya$^{3}$,
  Ruimin Dai$^{1}$,\\
  \textbf{Xiaoyan Han}$^{1}$,
  \textbf{Yanjie Fu}$^{4}$,
  \textbf{Dakuo Wang}$^{5}$,
  \textbf{Kunpeng Liu}$^{1}$\thanks{Corresponding author.}\\[3pt]
  $^{1}$Clemson University,\quad
  $^{2}$Rochester Institute of Technology,\quad
  $^{3}$Accenture (United States),\\
  $^{4}$Arizona State University,\quad
  $^{5}$Northeastern University\\[3pt]
  \texttt{jinghaz@clemson.edu, frmvcs@rit.edu, kunpenl@clemson.edu}
}

\begin{document}
\raggedbottom
\hbadness=10000
\vbadness=10000

\maketitle
\begin{abstract}
Reinforcement learning (RL) is effective in enhancing the accuracy of large language models in complex reasoning tasks.
Existing RL policy optimization frameworks rely on final-answer correctness as feedback signals and rarely capture the internal logical structure of the reasoning process. Consequently, the models would generate fluent and semantically relevant responses but logically inconsistent, structurally erratic, or redundant.
To this end, we propose \textbf{StaRPO}, a stability-augmented reinforcement learning framework that explicitly incorporates reasoning stability into the optimization objective. Our StaRPO decomposes stability into two computable lightweight metrics: the \textit{Autocorrelation Function} (ACF) to evaluate local step-to-step coherence, and \textit{Path Efficiency} (PE) to evaluate global goal-directedness of the reasoning trajectory. These stability rewards are combined with task rewards to provide complementary and process-aware feedback.
We validate the effectiveness of using ACF and PE rewards by showing their correlation with logic errors on two backbone models.
Experiments on four reasoning benchmarks show that StaRPO consistently outperforms compared baselines and can enhance both final-answer accuracy and logical stability.
\end{abstract}

\section{Introduction}
Recent advances in large language models (LLMs) for reasoning-intensive tasks demonstrate remarkable effectiveness. In particular, reinforcement learning (RL) has significantly raised the performance ceiling of LLMs on complex problem-solving benchmarks~\cite{guo2025deepseek,team2024qwen2,liu2024deepseek,pennino2025reasoning,bandyopadhyay2025thinking} with appropriate optimization approaches. 

However, reasoning is not merely about producing a correct final answer. From the perspective of human users, ideal reasoning should consist of three components: an accurate final result in terms of completing the task; a logically sound reasoning process to support the explainability and reliability of the result; and proper knowledge grounding for factual correction to improve the reliability and truthworthiness of the model's decisions and outputs.
Although final-answer accuracy and factual correctness can usually be clearly quantified and used as feedback signals in previous studies~\cite{zhang2024llm}, the logical soundness is a higher-level property describing the correctness of inferential relations between premises and conclusions~\cite{govier2010practical}. 
This is because logic does not have a quantitative formulation, but it is useful and widely recognized by human evaluators.
In complex real-world applications, such as multi-round decision-making, mathematical theorem proving, and medical consultations, LLMs often produce reasoning chains that, while fluent and contextually relevant, remain logically inconsistent or structurally erratic. Such flawed reasoning must be mitigated, as it risks providing users with spurious justifications or misleading conclusions.

\begin{figure}[htbp]
    \centering
    \includegraphics[width=0.8\linewidth]{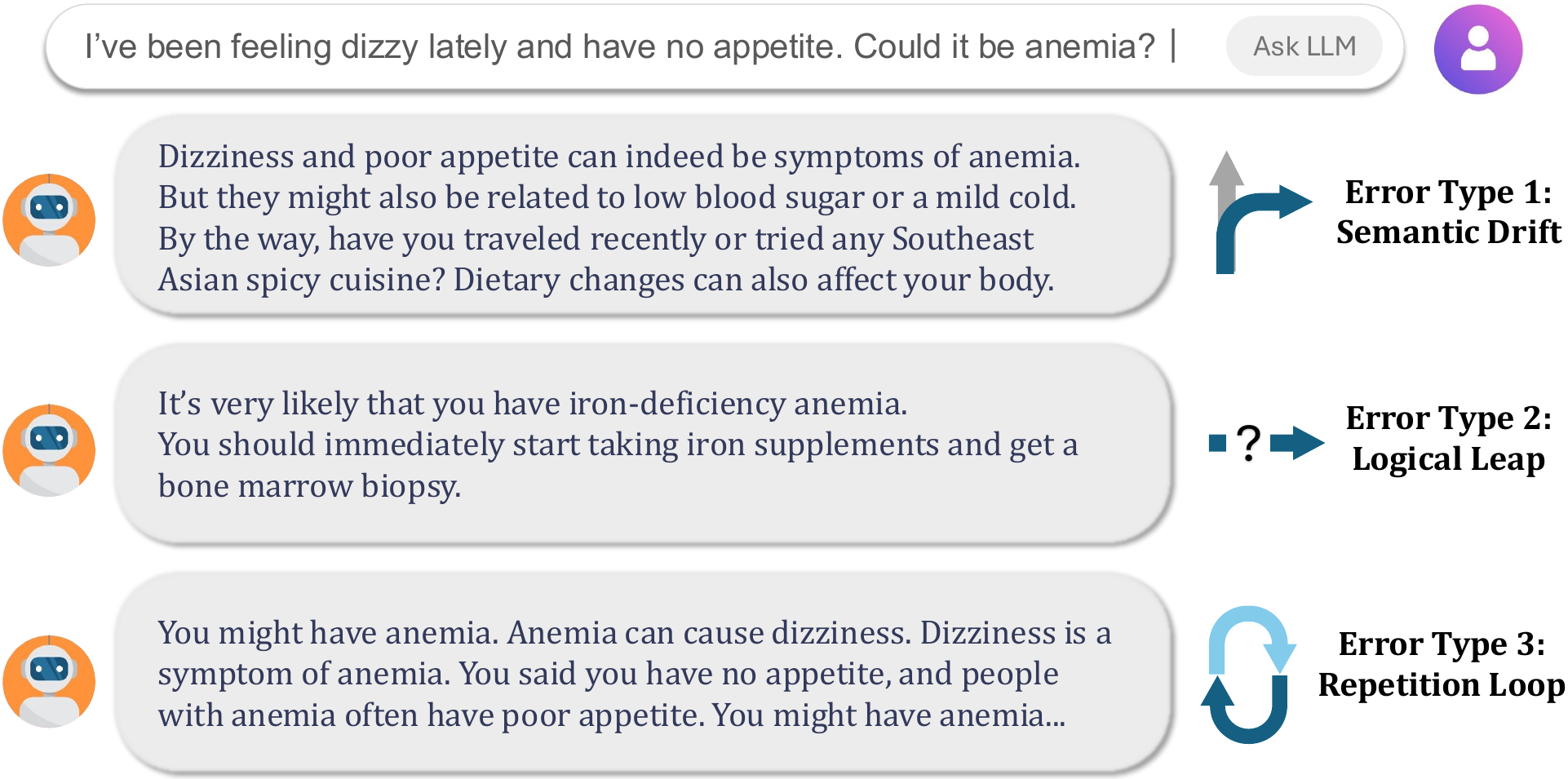}
    \caption{Example of three types of logical errors.}
    \label{fig:intro}
\vspace{-3ex}
\end{figure}

In Figure~\ref{fig:intro}, we provide three representative types of logical errors in a medical diagnosis scenario. The first response digresses from the core question to travel history. The second one skips the necessary step of inferring the disease from the given symptoms and directly goes to providing medical advice, while the third response falls into a redundant loop. 
All of these answers are semantically relevant to the query and expressed with a confident tone, yet none of them exhibit a coherent logical structure where conclusions are derived from preceding premises~\cite{george1995endorsement}.
In the representation space of potential responses, this phenomenon reflects that logically valid answers actually constitute only a small subset of the broader representation space of semantically plausible responses~\cite{yang2024enhancing}.
From a distributional perspective, generating logically valid responses is considerably more challenging than producing outputs that are merely semantically coherent~\cite{anderson2024semantic}.
These logical errors are difficult to detect in language models with confidence metrics (e.g., entropy or log-probabilities), even with available ground-truth answers, as they arise from deeper issues in how the model navigates and structures its internal reasoning trajectory~\cite{sun2023survey}. 

Although most result-based RL frameworks can significantly improve LLM reasoning performance~\cite{shao2024deepseekmath,feng2025group,lin2025cppo,zheng2025group}, their logical stability in terms of reasoning task are not satisfactory by solely relying on final-answer correctness as feedback signals.
The follow-up studies~\cite{zhang2025edge,chen2025ace} incorporate multiple types of rewards or signals for the reasoning process training, but still overlook the specific reasoning steps and abnormal logic within.

In this study, we aim to augment the logical correctness of the LLM reasoning process, i.e., the \textbf{logical stability} of reasoning. Logical stability contains three aspects: (1) \textbf{semantic coherence}: local step-to-step consistency in meaning, (2) \textbf{logical coherence}: the absence of discontinuities, leaps, or contradictions, and (3) \textbf{goal alignment}: maintaining a consistent trajectory toward the target answer. 
To this end, we build on the RL framework, with Group Relative Policy Optimization (GRPO) to incorporate additional trajectory-level signals and allow comparison of diverse reasoning behaviors.

\textit{Our target.} We target calculating and rewarding reasoning stability metrics effectively during training. Specifically, the stability is monitored from two perspectives: (1) \textbf{local coherence}, which ensures that each reasoning step logically follows from the previous one; (2) \textbf{global progression}, which ensures that the overall reasoning trajectory steadily advances toward a coherent final answer. Since reasoning stability is hard to quantify and is not a parametric property, we adopt RL as an alignment framework rather than using explicit supervision. Moreover, to avoid the high cost of using ``LLM-as-a-judge''~\cite{gu2024survey} for constructing feedback signals, we monitor the abnormal performance of reasoning trajectories in the embedding level as lightweight metrics for efficient computation.

\textit{Our method.} To this end, we introduce \textbf{StaRPO}, a \textbf{Sta}bility-Augmented \textbf{R}einforcement \textbf{P}olicy \textbf{O}ptimization framework. We quantify reasoning stability by decomposing it into two measurable metrics and converting them into rewards. These stability rewards are combined with task performance rewards to provide feedback for reasoning learning as supplementary. Specifically, we introduce the \textit{Autocorrelation Function (ACF)} and \textit{Path Efficiency (PE)} as two complementary metrics, where ACF evaluates the local stability by measuring directional changes at the embedding level, and PE evaluates the global stability by calculating the overall efficiency of the reasoning trajectory.

We conduct experiments to first evaluate the effectiveness of ACF and PE as indicators of logical stability. By examining the correlation between abnormal values of these metrics and three types of reasoning errors, we demonstrate that their combination provides a reliable signal of instability. We then conduct experiments with StaRPO on several reasoning benchmarks. Our experiments based on Qwen2.5-1.5B/7B-Instruct~\cite{team2024qwen2} models show that StaRPO consistently outperforms GRPO, where the generated reasoning steps are more accurate and reliable.

\section{Related Work}

\subsection{RL for Reasoning in LLMs}
Recent studies have demonstrated that RL can significantly enhance the reasoning capabilities of LLMs~\citep{liu2025reinforcement,zhang2025landscape} by taking the autoregressive generation as a sequential decision-making process and enabling the LLMs to optimize long-horizon behaviors through delayed reward signals~\citep{sun2023survey,zhang2025interplay}. 
Recent research has successfully applied RL across various domains, including mathematical reasoning, task decomposition, and agentic systems. These studies report significant and consistent gains in both output precision and overall task success rates~\citep{yue2025does,lin2025cppo}. 
Given that RL provides a robust framework for optimizing reasoning trajectories, we leverage it to enhance the logical consistency and inferential stability of LLM-generated outputs.

\subsection{Process-Level Rewards for Reasoning}
Moving beyond conventional outcome-based optimization, several studies have integrated auxiliary signals to refine the generation process. These process-oriented rewards encompass metrics such as response length, structural compliance, and information entropy. Furthermore, recent research has incorporated external verifier models or critics to provide a posterior evaluation of reasoning quality~\citep{wu2025lapo,yu2025rlpr,chen2025exploration,zhang2025r1}.
These process-level rewarding approaches enrich the signal space and correct some factorial error in sentence-level reasoning steps~\citep{zhang2025survey,cho2025breaking}, but cannot evaluate the logic evolution of the reasoning process.
However, most current process-level rewards are limited by their reliance on single metrics, task-specific heuristics, or post-hoc evaluations. Many of them depend on computationally expensive auxiliary models, making it difficult to provide stable and consistent supervision during the training phase. Consequently, these approaches frequently fail to provide the explicit supervision necessary to foster logical stability throughout the reinforcement learning training process.
 
\subsection{Embedding Directions and Logic Changes}
To quantify the logic differences among the potential generated texts in terms of a given sequence, existing studies use distances and directions of the embedding vectors as geometric measures of logical similarity and change~\citep{luus2021logic,nguyen2025representation}. 
This is because in distributed representations, relative positions and directional variations often reflect transitions between logic states~\citep{huang2022line,patil2023survey}.

In the context of autoregressive generation, the representations of the consecutive reasoning step form an embedding trajectory. Existing studies show that there is a correlation between the properties of trajectory (like the smoothness and consistency)~\citep{fang2016using,hosseiny2023review} and the semantic coherence and structural stability of generated text.
Conversely, abrupt or irregular directional changes within the trajectory are frequently associated with substantial semantic state  changes~\citep{wu2025ctrls,chen2025towards}. 
In the meantime, the logical errors frequently stem from unstable state transitions and anomalous semantic representations~\citep{tang2023neural,liu2025logic}.
To this end, we adopt the embedding of abnormal directional changes as signals to track potential logical instability during reasoning.
\section{Preliminary and Motivation}
In this paper, we take the autoregressive language model parameterized as a stochastic policy $\pi_\theta$.
Let $x$ denote a query (prompt) and $\mathcal{D}$ be the query set, for a response $y=(y_1,\ldots,y_{|y|})$ to query $x$, the policy likelihood is:
\begin{equation}
    \pi_\theta(y \mid x) = \prod_{t=1}^{|y|} \pi_\theta\left(y_t \mid x, y_{<t}\right),
\end{equation}
where $y_{<t}=(y_1,\ldots,y_{t-1})$ and $|y|$ is the response length. 
The reward model assigns a reward score $r(x,y) \in [0,1]$ to each query–response pair. In this paper, all expectations are taken over $x \sim \mathcal{D}$ and responses sampled from an ``old’’ policy $\pi_{\theta_{\text{old}}}$ unless stated otherwise.

\paragraph{Group Relative Policy Optimization (GRPO).}
GRPO~\citep{shao2024deepseekmath} estimates advantages through relative comparisons among multiple responses to the same query. Compared with PPO~\citep{schulman2017proximal}, GRPO enhances the scalability and robustness of training and is value-free, as it does not need a value function~\citep{guo2025deepseek}. GRPO computes the normalized advantages within a response group without KL regularization, whose optimization goal is:
\begin{equation}
J_{\text{GRPO}}(\theta) = \mathbb{E}_{x\sim\mathcal{D}, \{y_i\}_{i=1}^{G} \sim \pi_{\theta_{\text{old}}}} \Biggl[ \frac{1}{G} \sum_{i=1}^{G} \frac{1}{|y_i|} \sum_{t=1}^{|y_i|} 
\min \Bigl( w_{i,t}(\theta)\hat{A}_{i,t}, 
\text{clip}(w_{i,t}(\theta), 1-\epsilon, 1+\epsilon) \hat{A}_{i,t} \Bigr) \Biggr]
\end{equation}
where $G$ is the group size. The importance ratio $w_{i,t}(\theta)$ and advantage $\hat{A}_{i,t}$ of all token $y_{i,t} \in y_i$ are:
\begin{equation}
w_{i,t}(\theta)
= \frac{\pi_\theta(y_{i,t} \mid x, y_{i,<t})}
{\pi_{\theta_{\text{old}}}(y_{i,t} \mid x, y_{i,<t})},\quad
\hat{A}_{i,t}=\hat{A}_i
= \frac{r(x,y_i)-\text{mean}\big(\{r(x,y_j)\}_{j=1}^{G}\big)}
{\text{std}\big(\{r(x,y_j)\}_{j=1}^{G}\big)}.
\end{equation}

GRPO enhances the training stability and solves the reward sparsity by normalizing rewards among the group. It allows the policy to learn from relative comparison when the absolute reward signals are weak or missing. However, since the advantage $\hat{A_i}$ is computed solely with the reward of response, the signals are meaningful only when there is a sufficiently informative reward difference among the group. 
If all the responses in a group are wrong or similarly suboptimal, the relative advantage would not have a positive impact. 
Moreover, GRPO's optimization target is still based on the final output. Logically unstable reasoning trajectories are not identified and can not be used as signals to further distinguish the responses in a group.

\paragraph{Motivation.}
In the context of using RL training to enhance models' reasoning capability, a crucial question is \textit{how to take advantage of the reasoning trajectory to provide additional feedback signals?}
With these signals, the policy can improve the quality of reasoning paths while maintaining high final-answer accuracy. Besides, when the responses in one group have similar or equally-poor outcomes, these complementary signals can differ the responses and thus mitigate the limitation of the group-relative reward mechanism.

Given the potential effectiveness of these signals for policy optimization, another key question is: \textit{what kind of signal can effectively reflect the logical stability of a reasoning trajectory?} Meanwhile, considering the practical deployment in large-scale RL training, these signals should also be lightweight, easy to compute, and stable across different tasks and model scales, without relying on additional supervision or expensive external models. 
To explore these questions, we explore the signals that can indicate logic-level changes and design a corresponding mechanism to merge the signals into the RL optimization process.
\section{StaRPO: Stability-Augmented Reinforcement Policy Optimization}
In this section, we introduce StaRPO, a stability-augmented reinforcement policy optimization framework for improving reasoning reliability for LLMs.
We first define the logical-stable space and the reasoning trajectory's coherent evolution within logical constraints. Then, we introduce two complementary metrics to quantify local. With these metrics, we extend the GRPO framework by incorporating stability-augmented rewards into the optimization objective.  

\subsection{Logical-Stable Space}
\begin{wrapfigure}{l}{0.52\textwidth}
    \vspace{-0.5\baselineskip}
    \centering
    \includegraphics[width=0.5\textwidth]{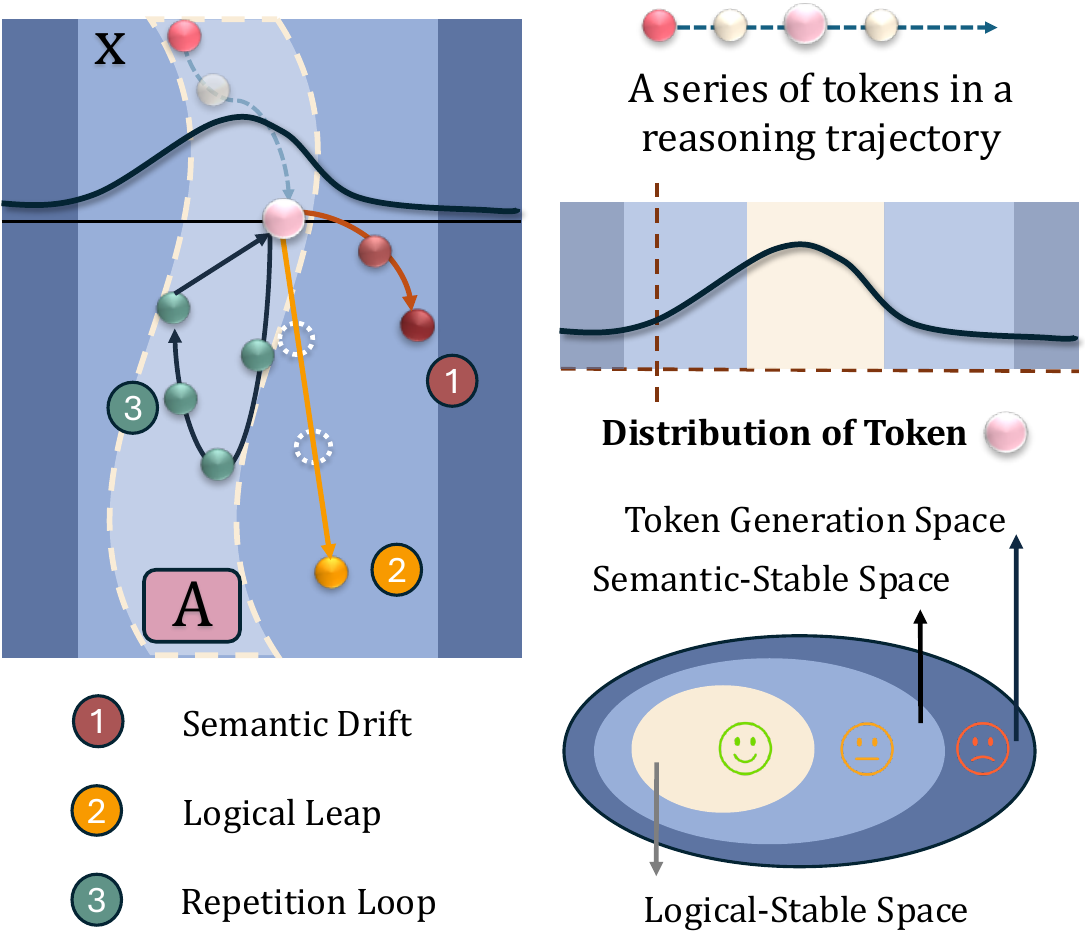}
    \caption{Logical-stable space in token generation space for a reasoning trajectory. From a query $x$, the LLM generates a sequence of token-level transitions targeting the correct answer region $A$ step-by-step. Incorrect logic may result in three types of errors.}
    \label{fig:stablespace}
    \vspace{-0.5\baselineskip}
\end{wrapfigure}

As shown in Figure~\ref{fig:stablespace}, at any decoding step $t$, given a query $x$ and the generated prefix $y_{<t}$, an autoregressive model can determine a conditional distribution $\pi_\theta(\cdot \mid x, y_{<t})$. For an ideal multi-step reasoning process, the model would sample one token per step from the conditional distribution. This sequence of token-level decisions is formulated as a series of state transitions in semantic space, which is used to form a reasoning trajectory from $x$ to a correct answer region $A$. 

The support of the token's conditional distribution constitutes the token generation space, denoted as $\mathcal{S}_t$.
This space characterizes all token choices that the model assigns a non-negligible probability under the current generation context. In $\mathcal{S}_t$, there exists a subset $\mathcal{S}_t^{\text{sem}} \subset \mathcal{S}_t$, which we refer to as the \textbf{semantic-stable space} $\mathcal{S}_t^{\text{sem}}$. As the working procedure of most well-trained LLMs, the tokens sampled from this region can make coherent, fluent, and semantically plausible text. 
However, as shown in Figure~\ref{fig:intro}, the output text from this semantic-stable space is acceptable at the language level, but does not necessarily guarantee logical stability. In the semantic-stable space $\mathcal{S}_t^{\text{sem}}$, there exists a much smaller and more restrictive subset $\mathcal{S}_t^{\text{log}} \subset \mathcal{S}_t^{\text{sem}}$, denoted as \textbf{logical-stable space}. In this space, the tokens not only maintain semantic coherence but also keep state transitions that are consistent with the logical structure of the task. If the generated sequence satisfies $y_t \in \mathcal{S}_t^{\text{log}}$ at every decoding step, the reasoning trajectory can progress continuously under logical constraints. At the textual level, each subsequent inference step builds upon its predecessors, progressing incrementally toward the final target answer $A$.

If at any step $t$ the generated token satisfies $y_t \notin \mathcal{S}_t^{\text{log}}$, the reasoning trajectory may shift from the intended direction and produce a logically wrong generation. The three types of logical errors occur when the trajectory goes into the semantic-stable space but not in the logical-stable subspace: (1) The red trajectory corresponds to semantic drift, that is, individual statements may remain fluent, the overall trajectory go wrong direction and not to a path to the reasoning target; (2) The yellow trajectory corresponds to logical leap, that is, the model skips several logically necessary intermediate states and leaps to a semantically far away region; (3) The green trajectory corresponds to repetition loop, that is, the generation lost direction and wonders within a local semantic neighborhood.



\subsection{Stability-Augmented Rewards}
In the last section, we characterized the reasoning process as an evolving token-level trajectory and analyzed how it can fail in logical instability. We now turn to the question of how such failures can be alleviated or prevented during RL training. 

In practice, it is hard to directly define and reward correct and stable logic even with an RL framework. Logical correctness is an abstract, high-level property that often depends on holistic human judgment and is hard to formalize or automatically evaluate at scale. In contrast, identifying unstable logical errors is much easier. Thus, our aim to introduce appropriate penalties in the reward signal when errors are highly likely exists and guides models back to a logical stable space. 

In addition, reasoning stability exists at both local and global levels. Local level emphasizes continuity and consistency, which requires continuous reasoning states to evolve smoothly and consistently without sudden leaps or big direction changes. Global level requires that the overall trajectory should make meaningful progress toward solving the target problem, instead of stalling or looping redundantly in the semantic space. Thus, we design our \model method from both local and global perspectives and the corresponding stability-augmented feedback signals to constrain and optimize the stability of model reasoning.

\subsubsection{Local: Stepwise Autocorrelation Reward}
To capture local level stability, we segment the LLM's reasoning generation into $K$ steps $\{Step_k\}_{k=1}^{K}$ and represent each step $Step_k$ by a semantic embedding vector $h_k$ which is the average of token-level hidden states within the step. We evaluate the evolution of reasoning as the embedding changes through steps.
We use the difference between consecutive step embeddings as a proxy for the local transition direction:
\begin{equation}
    \Delta h_k = h_k - h_{k-1}, \quad k=2,\ldots,K
\end{equation}
$\Delta h_k$ is expected to capture how the semantic state moves among reasoning steps.
A stable reasoning process should exhibit consistent directional changes across steps, whereas reversals and large directional deviations usually occur alongside unstable logical behavior, e.g., sudden leaps that skip necessary intermediate states. In contrast, persistently small turning angles over many steps can also be an abnormal sign, as it may reflect repetitive updates that do not introduce new information.

Intuitively, we adopt an autocorrelation-based metric to quantify the consistency between consecutive transition directions. Autocorrelation is originally a concept in time-series analysis to measure temporal dependence between adjacent observations in a sequence, and captures whether successive states evolve consistently over time. Here we draw an analogy by treating a reasoning trajectory as an ordered sequence of semantic state transitions. The consecutive transitions are analogous to adjacent observations in a time series.
Specifically, we define the stepwise autocorrelation function (\textbf{ACF}) as the cosine similarity between adjacent changes as:
\begin{equation}
\text{ACF}_k = \cos(\Delta h_k,\Delta h_{k-1})=\frac{\langle \Delta h_k,\Delta h_{k-1} \rangle}{\|\Delta h_k\| \cdot \|\Delta h_{k-1}\|}.
\end{equation}
Moderately positive $\text{ACF}_k$ values indicate locally consistent progression, and values that are too low (weak or negative alignment) or excessively high (near-identical updates repeated over steps) are often associated with unstable behaviors such as logical leaps or repetition loops.
Then, the ACF reward of a reasoning chain is the average over all steps as Eq.~\ref{eq: ACF}, which encourages coherent progression between steps and penalizes abrupt semantic jumps or redundancy.
\begin{equation}
\label{eq: ACF}
r_{\text{acf}} = \frac{1}{K-1}\sum_{k=2}^{K} \cos\big(\Delta h_k,\Delta h_{k-1}\big).
\end{equation}

\subsubsection{Global: Path Efficiency Reward} 
Simply using ACF to capture the local consistency between adjacent reasoning steps is not enough. A reasoning process could be locally consistent but fail globally, like wandering in semantic space or repeating without getting close to the solution. To address this limitation, we introduce a complementary metric that evaluates global stability from a goal-directed progression aspect.

A globally stable reasoning trajectory should transit efficiently from the initial state toward the final state without unnecessary detours. Based on this intuition, we quantify global progression by comparing the net displacement of the trajectory with its total path length.
Specifically, we define the net displacement $D$ to measure how far the reasoning state moves from the beginning to the end, and then the total path length $L$ can be calculated accordingly. 
This path length $L$ accumulates all intermediate semantic transitions along the trajectory. Then, the path efficiency $\eta$ is formulated in Eq.~\ref{eq: length}, where $\epsilon$ is a small constant for numerical stability.
\begin{equation}
\label{eq: length}
    D = \|h_K - h_1\|, \quad L = \sum_{k=2}^{K} \|h_k - h_{k-1}\|, \quad \eta = \frac{D}{L + \epsilon}, \quad \eta \in [0,1],
\end{equation}
A value $\eta \approx 1$ indicates a straight, goal-directed trajectory (stable reasoning), while $\eta \to 0$ reflects oscillations or loops (unstable reasoning).  
We use the path efficiency (\textbf{PE}) as the global stability reward in Eq.~\ref{eq: PE}. This reward explicitly encourages reasoning trajectories to make steady, goal-aligned progress toward the final answer, while penalizing locally fluent but globally inefficient trajectories.
\begin{equation}
\label{eq: PE}
r_{\text{pe}} = \frac{\|h_K - h_1\|}{\sum_{k=2}^{K} \|h_k - h_{k-1}\| + \epsilon}.
\end{equation}

\subsection{StaRPO Objective}
On top of the GRPO framework, we incorporate reasoning stability feedback into the policy optimization. 
For GRPO, the advantage of each response $y_i$ is computed based on its relative outcome-level reward within a group of responses to the same query. In \model, we extend this formulation by adding a stability-augmented reward for each reasoning trajectory as Eq.~\ref{eq: reward}, where $\lambda_{\text{acf}}, \lambda_{\text{pe}} \geq 0$ control the trade-off between task success and stability.
\begin{equation}
\label{eq: reward}
r_{\text{StaRPO}}(x, y_i)
= r(x, y_i)+ \lambda_{\text{acf}} \cdot r_{\text{acf}}+ \lambda_{\text{pe}} \cdot r_{\text{pe}},
\end{equation}
Accordingly, we compute group-relative advantages with the stability-augmented rewards. For a group of $G$ responses ${y_i}_{i=1}^{G}$ generated for the same query $x$, the advantage for response $y_i$ is:
\begin{equation}
A_i=\frac{r_{\text{StaRPO}}(x, y_i)-\text{mean}\big(\{r_{\text{StaRPO}}(x,y_j)\}_{j=1}^{G}\big)}
{\text{std}\big(\{r_{\text{StaRPO}}(x, y_j)\}_{j=1}^{G}\big)}.
\end{equation}
where all tokens within the same response share the same scalar advantage $A_i$. Ultimately, the overall objective function of StaRPO is formulated as:
\begin{equation}
J_{\text{StaRPO}}(\theta) = \mathbb{E}_{\substack{x \sim \mathcal{D} \\ \{y_i\}_{i=1}^{G} \sim \pi_{\theta_{\text{old}}}}} 
\Biggl[ \frac{1}{G} \sum_{i=1}^{G} \frac{1}{|y_i|} \sum_{t=1}^{|y_i|} \min \Bigl( w_{i,t}(\theta) A_i, \
\text{clip}\bigl(w_{i,t}(\theta), 1-\epsilon, 1+\epsilon\bigr) A_i \Bigr) \Biggr].
\end{equation}

\section{Experiments}

\subsection{Experiment Setup}
\paragraph{Evaluation Benchmarks and Metrics.}
We use four reasoning benchmarks to evaluate our \model, 
including: (1) GSM8K~\citep{cobbe2021training}, which contains grade-school math word problems testing multi-step arithmetic reasoning; (2) MATH~\citep{hendrycks2021measuring} contains competition-level math questions;
(3) GPQA-Diamond~\citep{rein2024gpqa}, a curated subset of GPQA, which contains 198 PhD-level science questions authored by domain experts in physics, chemistry, and biology; and (4) AIME24~\citep{mathai_aime24} contains 30 competition-style math problems covering arithmetic, algebra, and geometry from the American Invitational Mathematics Examination.
We use final-answer accuracy \textbf{(Result)} as well as the process accuracy \textbf{(Process)}, which measures the proportion of reasoning trajectories with no detected logical errors, as evaluation metrics.

\paragraph{Baselines.}
We compare our method with two types of baselines. The first category constrains model behavior by modifying RL training objectives, including (1) \textbf{GRPO}~\citep{shao2024deepseekmath}, which optimizes policies via group-relative outcome comparison without a value model; (2) \textbf{CPPO}~\citep{lin2025cppo}, which improves GRPO efficiency by pruning low-advantage completions during training; and (3) \textbf{$\boldsymbol{\lambda}$-GRPO}~\citep{wang2025lambda}, which mitigates GRPO's length bias by learning adaptive token-level weights without introducing additional process-level supervision.
Another one category consists of methods that directly supervise or intervene in the reasoning process using process-level signals, including (4) \textbf{Entropy}-based control~\citep{zhang2025entropy}, which guides reasoning depth based on model uncertainty estimates; and (5) LLM-\textbf{Planner}~\citep{song2023llm}, which leverages external planning modules to explicitly guide intermediate decision making during task execution.

\paragraph{Implementation Details.}
All RL-based methods are implemented following the previous studies~\citep{lin2025cppo,wang2025lambda} with the same training configuration to ensure fair comparison. Unless otherwise specified, the group size is fixed to $8$, and all models are trained for two RL iterations with a KL regularization coefficient $\beta=0.04$. We conduct our experiment on \texttt{Qwen2.5-Instruct} in two sizes, 1.5B and 7B~\citep{team2024qwen2}. The sampling temperature is set to $1.0$, except for GSM8K as $0.9$, following common practice~\citep{lin2025cppo}. The learning rate is set to $1e-6$.
For \model, we estimate the empirical distributions of ACF and PE metrics using the first 100 generated reasoning trajectories for each benchmark, and use the corresponding tail regions as the initial abnormality ranges for both metrics. 
All experiments are conducted on four NVIDIA H200 80G GPUs.

\begin{table*}[ht]
\centering
\small
\setlength{\tabcolsep}{3.5pt}
\vspace{-2mm}
\resizebox{\textwidth}{!}{%
\begin{tabular}{l l rr rr rr rr rr}
\toprule
\textbf{Model} & \textbf{Method}
& \multicolumn{2}{c}{\textbf{GSM8K}}
& \multicolumn{2}{c}{\textbf{MATH}}
& \multicolumn{2}{c}{\textbf{GPQA}}
& \multicolumn{2}{c}{\textbf{AIME}}
& \multicolumn{2}{c}{\textbf{Avg.}} \\
\cmidrule(lr){3-4} \cmidrule(lr){5-6} \cmidrule(lr){7-8} \cmidrule(lr){9-10} \cmidrule(lr){11-12}
& & Result & Process
& Result & Process
& Result & Process
& Result & Process
& Result & Process \\
\midrule

\multirow{7}{*}{Qwen 1.5B}
& Original & 55.72 & 52.39 & 47.73 & 43.94 & 19.19 & 16.67 & 3.33 & 3.33 & 31.49 & 29.08 \\
\cmidrule(lr){2-12}
& CPPO     & 76.80 & 73.62 & \cellcolor{second}54.55 & 52.08 & \cellcolor{second}34.85 & 28.79 & \cellcolor{best}\textbf{13.33} & \cellcolor{best}\textbf{10.00} & \cellcolor{second}44.88 & 41.12 \\
& $\lambda$--GRPO & \cellcolor{second}78.17 & \cellcolor{second}76.95 & 53.98 & \cellcolor{second}52.65 & 33.33 & \cellcolor{second}30.81 & 6.67 & 6.67 & 43.04 & \cellcolor{second}41.77 \\
& GRPO     & 73.92 & 72.02 & 53.60 & 49.62 & 26.77 & 23.23 & \cellcolor{second}10.00 & \cellcolor{best}\textbf{10.00} & 41.07 & 38.72 \\
\cmidrule(lr){2-12}
& Entropy  & 65.05 & 64.29 & 52.65 & 50.57 & 25.25 & 23.23 & 6.67 & 6.67 & 37.41 & 36.20 \\
& Planner  & 60.35 & 58.83 & 49.81 & 47.35 & 22.73 & 19.70 & 0.00 & 0.00 & 33.22 & 31.47 \\
\cmidrule(lr){2-12}
& StaRPO   & \cellcolor{best}\textbf{79.68}$^\dagger$ & \cellcolor{best}\textbf{78.54}$^\dagger$ &
            \cellcolor{best}\textbf{56.25}$^\dagger$ & \cellcolor{best}\textbf{56.06}$^\dagger$ &
            \cellcolor{best}\textbf{35.35}$^\dagger$ & \cellcolor{best}\textbf{34.85}$^\dagger$ &
            \cellcolor{second}10.00 & \cellcolor{best}\textbf{10.00} &
            \cellcolor{best}\textbf{45.32}$^\dagger$ & \cellcolor{best}\textbf{44.86}$^\dagger$ \\

\midrule

\multirow{7}{*}{Qwen 7B}
& Original & 89.46 & 88.63 & 53.60 & 52.27 & 28.79 & 26.77 & 10.00 & 10.00 & 45.46 & 44.42 \\
\cmidrule(lr){2-12}
& CPPO     & \cellcolor{second}90.52 & 88.93 & 74.43 & 71.97 & \cellcolor{best}\textbf{38.89} & \cellcolor{second}35.35 & \cellcolor{second}23.33 & \cellcolor{second}20.00 & \cellcolor{second}56.79 & \cellcolor{second}54.06 \\
& $\lambda$--GRPO & 90.90 & 89.08 & \cellcolor{second}75.38 & \cellcolor{second}74.62 & 37.88 & 34.85 & 16.67 & 13.33 & 55.21 & 52.97 \\
& GRPO     & 90.30 & \cellcolor{second}89.39 & 74.24 & 73.11 & 35.86 & 33.33 & \cellcolor{second}23.33 & 16.67 & 55.10 & 53.13 \\
\cmidrule(lr){2-12}
& Entropy  & 90.37 & 89.23 & 67.61 & 65.72 & 30.30 & 29.80 & 13.33 & 13.33 & 50.40 & 49.52 \\
& Planner  & 89.76 & 89.08 & 52.08 & 50.57 & 28.79 & 24.24 & 6.67 & 6.67 & 44.33 & 42.64 \\
\cmidrule(lr){2-12}
& StaRPO   & \cellcolor{best}\textbf{91.66}$^\dagger$ & \cellcolor{best}\textbf{91.43}$^\dagger$ &
            \cellcolor{best}\textbf{77.46}$^\dagger$ & \cellcolor{best}\textbf{76.89}$^\dagger$ &
            \cellcolor{second}38.38$^\dagger$ & \cellcolor{best}\textbf{37.88}$^\dagger$ &
            \cellcolor{best}\textbf{26.67} & \cellcolor{best}\textbf{23.33} &
            \cellcolor{best}\textbf{57.71}$^\dagger$ & \cellcolor{best}\textbf{57.38}$^\dagger$ \\

\bottomrule
\end{tabular}%
}
\vspace{-2mm}
\caption{Performance comparison across four tasks. The best and second-best performance is set in \colorbox{best}{dark} and \colorbox{second}{light} blue. Superscript $\dagger$ denotes significant improvements with paired t-test at $p<0.05$ over GRPO.}
\label{tab:main-results}
\vspace{-3mm}
\end{table*}

\subsection{Overall Performance}
We first conduct standard training and evaluation across four benchmarks for both final-answer accuracy and process accuracy. As shown in Table~\ref{tab:main-results}, we observe that \model consistently achieves the best or second-best performance across all datasets and backbone models, especially in knowledge-intensive tasks (GPQA) and long-horizon reasoning tasks (MATH). 
Specifically, \model gains more improvements in process accuracy. 
For example, on Qwen-1.5B, \model improves process accuracy over GRPO by $+11.62\%$ on GPQA, which is $3.04\%$ higher than the improvement in final-answer accuracy.
This observation suggests that the stability-augmented rewards primarily enhance the logical quality of the reasoning process rather than solely increasing the chance of producing a correct answer. About the compared baselines, although GRPO and those that modify RL optimization mechanisms (CPPO and $\lambda$-GRPO) can improve final-answer accuracy by providing stable and efficient optimization signals, these methods cannot provide stable enhancement in process accuracy as our \model. The average improvement for CPPO and $\lambda$-GRPO over GRPO in process accuracy is $+2.4\%$ and $+3.05\%$, respectively, while our \model provides $+6.14\%$ on Qwen-1.5B. Besides, the process accuracy of CPPO and $\lambda$-GRPO even drops based on the 7B model on GSM8K.
Entropy-based control and planning methods achieve smaller improvements, and we can see degradation on complex tasks. This phenomenon suggests that relying on uncertainty signals or an external planning module without optimization policies is difficult to provide stable and scalable improvements in reasoning quality. In contrast, \model directly incorporates stability-augmented signals into the RL reward, enabling consistent and robust enhancements in both process and final-answer accuracy.

\begin{figure}[h]
\centering
\begin{tcolorbox}[starpoOneBox, width=\linewidth, top=5pt, bottom=5pt, left=8pt, right=8pt]
{\footnotesize
Question: Use numbers and basic arithmetic operations (+ - * /) to obtain 24. \\
Example 1 (Semantic Drift):\
Input: 5 5 5 9 \
Model Output: 5 * 5 = 25, 25 / 9 = 2.7778, 2.7778 / 9 = 0.309 \ldots \\
Example 2 (Logical Leap):\
Input: 5\ 5\ 5\ 9\
Model Output: 19 + 5 = 24 \\
Example 3 (Repetition Loop):\
Input: 8 4 3 1 \
Model Output: 8 / 4 = 2, 2 * 4 = 8, 8 / 4 = 2, 2 * 4 = 8 \ldots
}
\end{tcolorbox}
\caption{Example of logical reasoning errors on \textit{Game of 24}.}
\label{fig:cot-failure-examples}
\end{figure}
\subsection{Correlation between Logical Errors and Stability Metrics}
In this experiment, we investigate whether the abnormal values of the two proposed metrics, ACF and PE, are significantly correlated with the three types of logical errors (semantic drift, logical leap, and repetition loop) as introduced in Figure~\ref{fig:intro}. 
We conduct this experiment on Task \textit{Game of 24}~\citep{NEURIPS2023_271db992}, which 
requires the model to obtain the target value 24 by applying basic arithmetic operations to a given set of numbers (\textit{``Use numbers and basic arithmetic operations (+ – × ÷) to obtain 24. Example Input: 2, 4, 6, 8''}). 
The reasons of using this task are: (1) The reasoning process is highly transparent. For example, when we take 4 and 8 to achieve a 12 as ``4 + 8 = 12'', and then we would have 2, 6 from the original input set, and the newly generated 12 to continue to get 24, e.g., ``2 $\times$ 6 + 12 = 24''. The procedure can be naturally segmented into a sequence of steps; 
(2) The objective is clear, with well-defined evaluation criteria (whether the result equals 24 with all numbers and operations used correctly); 
and (3) No world knowledge or textual context is involved as the errors arise purely from the reasoning process, not from language or factual gaps. 
For each puzzle, the model generates an answer in eight reasoning steps, then we use \texttt{GPT-4o-mini}~\citep{gpt4omini} to annotate whether it exhibits any of the three error types, as shown in Figure~\ref{fig:cot-failure-examples}.

For each reasoning chain, we calculate its ACF and PE values with different rules for \textit{abnormal} values:
for ACF, we consider both abnormally low values (indicating inconsistent transition directions) and abnormally high values (often associated with rigid or repetitive progression), and for PE, we focus on significantly low values that reflect inefficient global trajectories. Specifically, we take the bottom or top $15.87\%$ tails of the value distribution as abnormal following the $3\sigma$ rule under a Gaussian assumption. To quantify the association between stability metrics and logical errors, we perform significance tests for each error type and examine whether samples with abnormal metric values exhibit higher error occurrence rates. We employ both two-sample $t$-tests and \textit{Mann--Whitney} tests to ensure our conclusions are robust under different distributional assumptions. We adopt a significance level of $\alpha=0.05$, the null hypothesis assumes no association between stability metric anomalies and logical errors, and is rejected when $p<0.05$.

\begin{table}[t]
\vspace{-2mm}
\centering
\resizebox{0.65\linewidth}{!}{
\begin{tabular}{lcccc}
\hline
\toprule
\textbf{Error Type} & \textbf{ACF Low} & \textbf{ACF High} & \textbf{PE (t-test)} & \textbf{PE (M-W)} \\
\midrule
Semantic Drift & {0.0217*} & {0.0207*} & {0.0358*} & 0.1363 \\
Logical Leap   & {0.0063**} & {0.0011**} & {0.0251*} & {0.0483*} \\
Repetition     & 0.2158                  & {0.0297*} & {0.0408*} & {0.0379*} \\
\bottomrule
\end{tabular}
}
\caption{Significance test between reasoning errors and stability metrics. Superscript * and ** denotes $p < 0.05$ and $p < 0.01$, respectively.}
\label{tab:combined-stability-significance}
\end{table}

\begin{figure*}[t]
    \centering
    \vspace{-2mm}
    \includegraphics[width=0.93\linewidth]{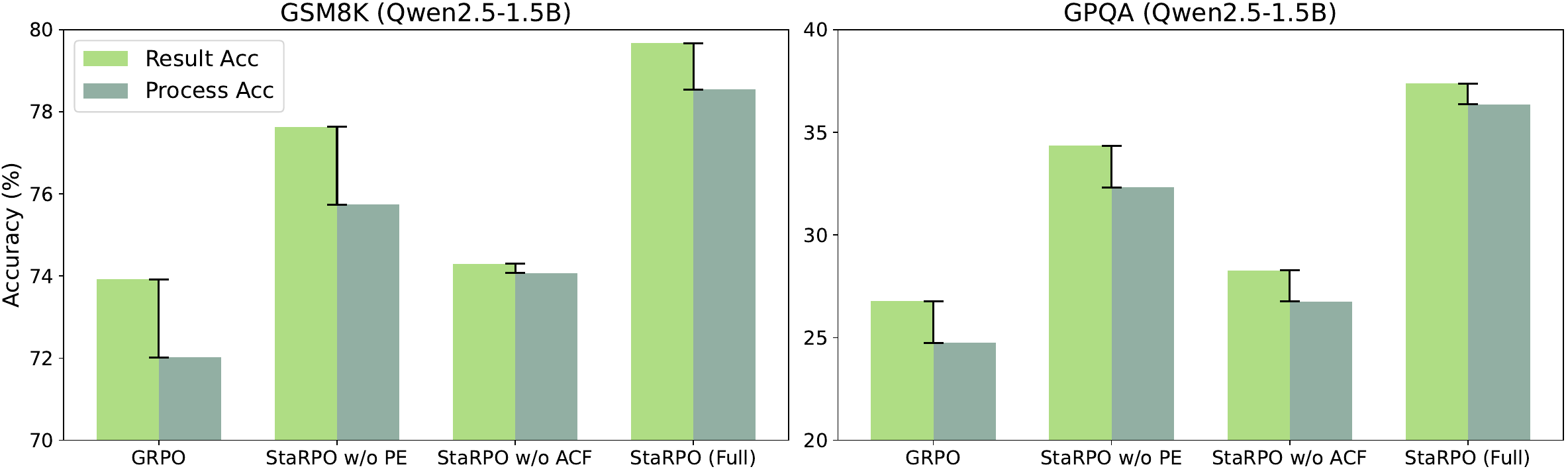}
    \caption{Ablation studies with result and process accuracy across different model variants on two datasets.}
    \label{fig:ablation}
\end{figure*}

As shown in Table~\ref{tab:combined-stability-significance}, both ACF and PE exhibit statistically significant associations with logical errors across multiple settings. For semantic drift, both low and high ACF values are significantly associated with error occurrence. This phenomenon indicates that drift may arise from inconsistent local transitions or overly rigid progression. The association with PE is significant under the $t$-test but not under the \textit{Mann–Whitney} test, which indicates a weaker effect. For the logical leap, all metrics show consistent significance across tests, which indicates that the logical leap is closely related to both local transition instability and global inefficiency. For repetition, high ACF and PE show a significant association.
Overall, the null hypothesis of no association is rejected. The abnormal ACF and PE values are significantly associated with reasoning instability errors.

\subsection{Ablation Study}
We conduct ablation studies on GSM8K and GPQA using Qwen 2.5-1.5B-Instruct to analyze the individual contributions of ACF and PE reward, respectively. Specifically, we compare four variants: (1) Original \textit{GRPO}; (2) Our \model without PE reward (\textit{StaRPO w/o PE}); (3) Our \model without ACF reward (\textit{StaRPO w/o ACF}); and (4) \textit{StaRPO (Full)}, which contains both ACF and PE. For each method, we report both final result accuracy and process accuracy.

As shown in Figure~\ref{fig:ablation}, in terms of reasoning performance, both ACF and PE rewards contribute positively across two datasets.
Removing either ACF or PE results in a noticeable decrease in both result and process accuracy, suggesting that each component captures a distinct aspect of reasoning quality for constraining reasoning stability during RL training.
In particular, removing ACF results in a substantial drop in process accuracy, which indicates the importance of local transition consistency for maintaining structurally reliable reasoning paths.
The unstable local semantic transitions can easily propagate and disrupt subsequent reasoning steps, even when the overall direction remains plausible. 
By contrast, combining ACF and PE consistently achieves the best performance across both metrics, indicating that local and global stability signals are complementary rather than redundant. 
\section{Conclusion}
In this paper, we introduce StaRPO, a stability-augmented reinforcement policy optimization framework to enhance the logic stability of the LLM reasoning process.
To capture the internal structure of reasoning trajectories, we focus on the notion of logical stability and decompose it into complementary local and global components.
Two lightweight stability metrics, ACF and PE, are designed to capture step-wise consistency and goal-directed progression. 
Our experiments across four reasoning benchmarks demonstrate the effectiveness of \model by comparing with existing strong baselines.


\bibliographystyle{plainnat}
\bibliography{references}


\end{document}